# Modelling the emergence of open-ended cultural evolution

**James Winters**[1,*] **Mathieu Charbonneau**[2]

[1]Centre for Culture and Evolution, Brunel University of London

[2]Africa Institute for Research in Economics and Social Sciences, Université Mohammed VI Polytechnique

[*]Corresponding author

January 05, 2026

**Abstract**

Humans stand alone in terms of their potential to collectively and cumulatively change their culture in an open-ended manner. This open-endedness provides societies with the ability to continually expand their resources and to increase their capacity to store, transmit and process information at a collective-level. Here, we propose that the production of resources arises from the interaction between cultural systems (a society's repertoire of interdependent techniques, artifacts, norms and knowledge) and search spaces (an ensemble of needs, problems and goals facing a society). Starting from this premise we develop a macro-level model wherein both cultural systems and search spaces are subject to evolutionary dynamics. By manipulating the extent to which these dynamics are characterised by stochastic or selection-like processes, we demonstrate that open-ended growth is extremely rare, historically contingent and only possible when cultural systems and search spaces co-evolve. Here, stochastic factors must be strong enough to continually perturb the dynamics into a far-from-equilibrium state, whereas selection-like factors help maintain effectiveness and ensure the sustained production of resources. Only when this co-evolutionary dynamic maintains effective cultural systems, supports the ongoing expansion of the search space and leads to an increased provision of resources do we observe open-ended cultural evolution.

## 1 Introduction

Over the last 300,000 years, our species has improved, diversified, and complexified the range of cultural traditions available to us, allowing human societies to collectively address various needs and goals as well as tackle diverse challenges and problems [1–4]. Even in their simplest guises, human cultures constitute complex systems of interdependent techniques, artifacts, norms and knowledge [5–8] that are collectively shared, culturally transmitted and cumulatively evolving [3,9–15]. Cultural systems are powerful in this respect because humans are able to collectively adapt [16,17] and discover highly optimized solutions for a wide variety of social and ecological niches [18]. What makes human culture special, however, is that it falls under a class of dynamics exhibiting *open-ended evolution* [19–33]: evolutionary processes capable of continually generating novelty and undergoing sustained, long-term growth in complexity. It is this open-ended capacity that ultimately underpins our ability to extract and harness diverse sources of energy, to access and manufacture novel materials, and to transmit and process information at vast scales [2,22,33–35].

Explaining why and how human culture is uniquely open-ended remains an unresolved challenge of the biological, social and cognitive sciences [22–24,26,36]. Computational models offer a particularly powerful tool for theory-building in this regard [37–41] and figure prominently in the literature on the evolution of culture [31,42–47]. However, as it currently stands, a strict separation exists between models focusing on the adaptive nature of cultural traits (e.g., [44,48–54]) and models focusing on the open-ended evolution of cultural systems (e.g., [43,45,55,56]). Adaptive approaches often employ search spaces [52] (also see fitness landscapes [57–59]) and model cultural evolution as a process of cumulative optimization (see [15,22,31] for similar perspectives). In these models, cultural traits exist in a bounded state space where different trait configurations map onto varying fitness or payoff values. A population then searches within this space by modifying, evaluating and selecting cultural traits based on maximising (or minimising) an objective function (e.g., optimizing an arrowhead to improve hunting [51,53]). Yet, despite the utility of such models for investigating evolutionary dynamics under the constraints of rich, high-dimensional search spaces (e.g., [49,51,52]), the bounded landscapes used are of limited applicability to questions of open-endedness [37,59].

Meanwhile, models of open-ended evolution tend to abstract away from explicitly representing search spaces and are instead concerned with the cumulative and combinatorial dynamics of cultural systems [43,45,55,56,60–62]. The explanatory goal here is to identify the conditions in which collections of cultural traits undergo long-term growth in complexity and/or diversity [43]. Yet, in omitting search spaces, these approaches effectively ignore an important source of constraints on the structure and function of cultural systems. This has led to various simplifying assumptions about cultural traits, their utility, and the costs and benefits of complexity. One pervasive assumption, which is present to varying degrees in all models surveyed, is that cumulative improvements only arise from the selection of novel cultural traits [45,55,56,62,63]. Loss, in cases where it is explicitly modelled, tends to be viewed as neutral or deleterious: cultural traits are either removed through stochastic perturbations (e.g., [64]) or as a result of selecting out low utility traits (e.g., [62]). Under such assumptions, a cultural system will inevitably experience open-ended growth so long as the mechanisms of generating and adopting novel cultural traits overcome any limiting mechanisms of loss.

A clear implication here is that existing models implicitly build-in a bias towards open-endedness. All that is required is some probability of generating and then selecting useful cultural traits. Still, even if we grant that this is a convenient simplification, it is not immediately obvious why the addition of cultural traits is the only pathway for discovering improvements. On the contrary, if we assume that cultural systems are adapting to the ecological and socio-economic demands of a search space [2,65], then a plausible adaptive response could be for a society to actively maintain or simplify its current cultural system (for similar arguments see [31,66,67]). Missing from these models is any formal or computational representation of the interactions between cultural systems and search spaces. Not only does this fail to capture how cultural systems adapt to an underlying search space, it also ignores the possibility that search spaces dynamically change in response to the cultural systems of a society.

To address these limitations, we develop a computational model that incorporates cultural systems and search spaces using a minimal set of processes and assumptions. Simplifying, our model abstracts away from individual-level interactions found in agent-based models [41], and models cultural evolution at the macro-level [43]. In our model, the dynamics of cultural evolution are conceived of as a co-evolutionary relationship between two fundamental processes: one which changes the structure of *cultural systems* (the repertoire of interacting techniques, artifacts, norms and knowledge available to a population) and another which changes the structure of *search spaces* (the ensemble needs, problems and goals facing a society at a given point in time). By modelling cultural evolution in this way, our model makes four contributions that link together cultural systems and search spaces with their potential for open-ended evolution.

First, we explicitly represent search spaces alongside the cultural systems of societies. Unlike previous models, where search spaces are either bounded (e.g., [50,52,53]), black boxed in favour of intrinsic utility/fitness values (e.g., [45,55,62]) or ignored altogether (e.g., [43,46,68]), we assume cultural systems and search spaces are structured. This provides us with a way of modelling the structure of cultural systems and search spaces as varying both in terms of their mutual fit as well as their complexity. In the context of our model, search spaces serve as adaptive targets that dynamically influence the adoption of changes to a cultural system. These changes include modifications (that maintain complexity), simplifications (that decrease complexity) and expansions (that increase complexity). Importantly, this recognises that the structure of search spaces is not inherently geared towards progressive increases in complexity. Instead, search spaces can impose pressures to maintain or even simplify the complexity of cultural systems.

Second, we propose that cultural systems *and* search spaces are subject to evolutionary dynamics. Needs, problems and goals are not static and invariant properties of the world, but rather form a dynamically shifting landscape that influences the direction of cultural evolution. Although it is relatively uncontroversial to model cultural systems as evolving entities [8,43,69,70], we consider parallel dynamics are at play in changing the search spaces of a society [32,33,71]. Extending the logic of cultural evolutionary theory to search spaces naturally follows if we think of needs, problems and goals as a form of information that is socially transmitted [72]. In aggregate, this information determines what constitutes an effective cultural system, and serves as an adaptive landscape to explore and exploit. Not only does our approach build upon the well-established notion that needs, problems and goals are cognitive phenomena [71,73–75], it also enables us to model search spaces as co-evolving with the cultural systems of a society.

Third, co-evolutionary dynamics in our model follow a two-step process of generating and then adopting changes to both cultural systems and search spaces. Crucially, this allows us to model co-evolution in terms of stochastic and deterministic factors. Stochastic factors, which are analogous to mutation and drift in biological evolution [76,77], assume random changes shape the evolutionary trajectory of cultural systems and search spaces. In a purely stochastic scenario, cultural systems and search spaces are maximally decoupled — the adoption of changes in one does not causally influence the adoption of changes in the other. Deterministic factors, which are analogous to selection dynamics in biology [78] or social learning biases [79] and selective filters [70] in culture, evaluate and adopt changes based on effectiveness: cultural systems are refined to more effectively exploit search spaces, while search spaces are restructured to more effectively align with the cultural capabilities of a society.

Fourth, in order to sustain the costs of changing both cultural systems and search spaces, societies need to produce and then allocate resources at each generation to these two processes. Resource constraints are a feature of any system subject to computational and energy demands. Human societies are no different and must therefore produce, consume and allocate resources in order to maintain their activities [34,35,80–82]. Differences in the ability to produce resources lead to differences in the capabilities of societies to generate and adopt changes to cultural systems and search spaces. Failure to procure sufficient resources is commonly associated with cultural stagnation [83] and societal collapse [84,85], whereas surplus resources are critical for societies to overcome the costs of increased complexity [67], to engage in innovation [82] and to maintain long-term cumulative growth [81,86]. A key feature of our model is that resources can theoretically grow without limit so long as societies expand their search space and maintain effective cultural systems.

Open-ended evolution in this context corresponds to a process of continually increasing the complexity of both cultural systems *and* search spaces. The challenge facing societies, then, is to simultaneously maintain effective cultural systems and engage in the ongoing expansion of the search space. Here, we assume more complex search spaces have a greater potential for producing resources than simpler ones. Yet, to realise this potential, societies need to also adopt improvements to their cultural systems. An inability to find effective outcomes is penalised by assuming the cost of maintaining cultural systems scales with their complexity. This represents the observation that supporting complex systems requires overcoming increasingly sophisticated problems associated with the production and consumption of resources [67,81,86,87]. We argue it is this co-evolutionary interaction between maintaining effective cultural systems, the ongoing expansion of the search space and the increased provision of resources that underpins the open-ended evolution of human culture.

In the following, we manipulate the extent to which the co-evolutionary dynamics between cultural systems and search spaces are stochastic or deterministic and examine in which conditions open-ended evolution emerges. Our findings suggest that the balance between stochastic and deterministic factors constitutes a fundamental limiting or enabling factor on the long-term sustainability of co-evolutionary dynamics. In particular, we show that in the vast majority of cases the co-evolutionary dynamics end up exhausting the available resources and hit an absorbing barrier [37,88]: a boundary condition on the ability of societies to generate and adopt changes to their cultural systems and search spaces. Open-ended growth is comparatively rarer and only emerges when both cultural systems and search spaces are shaped by sufficiently powerful selection pressures.

## 2 Model description: bitw0r1d

Bitw0r1d is a general framework for simulating cultural evolution. Adopting a macroscopic approach, which aims to explain long-term patterns of change using a minimal set of endogenous macro-level properties [43,61], we model the dynamics of cultural evolution in terms of two causally interacting processes: one that changes the structure of cultural systems ($\mathbb{C}$) and another that changes the structure of the search space ($\mathbb{S}$) (see Figure 1).

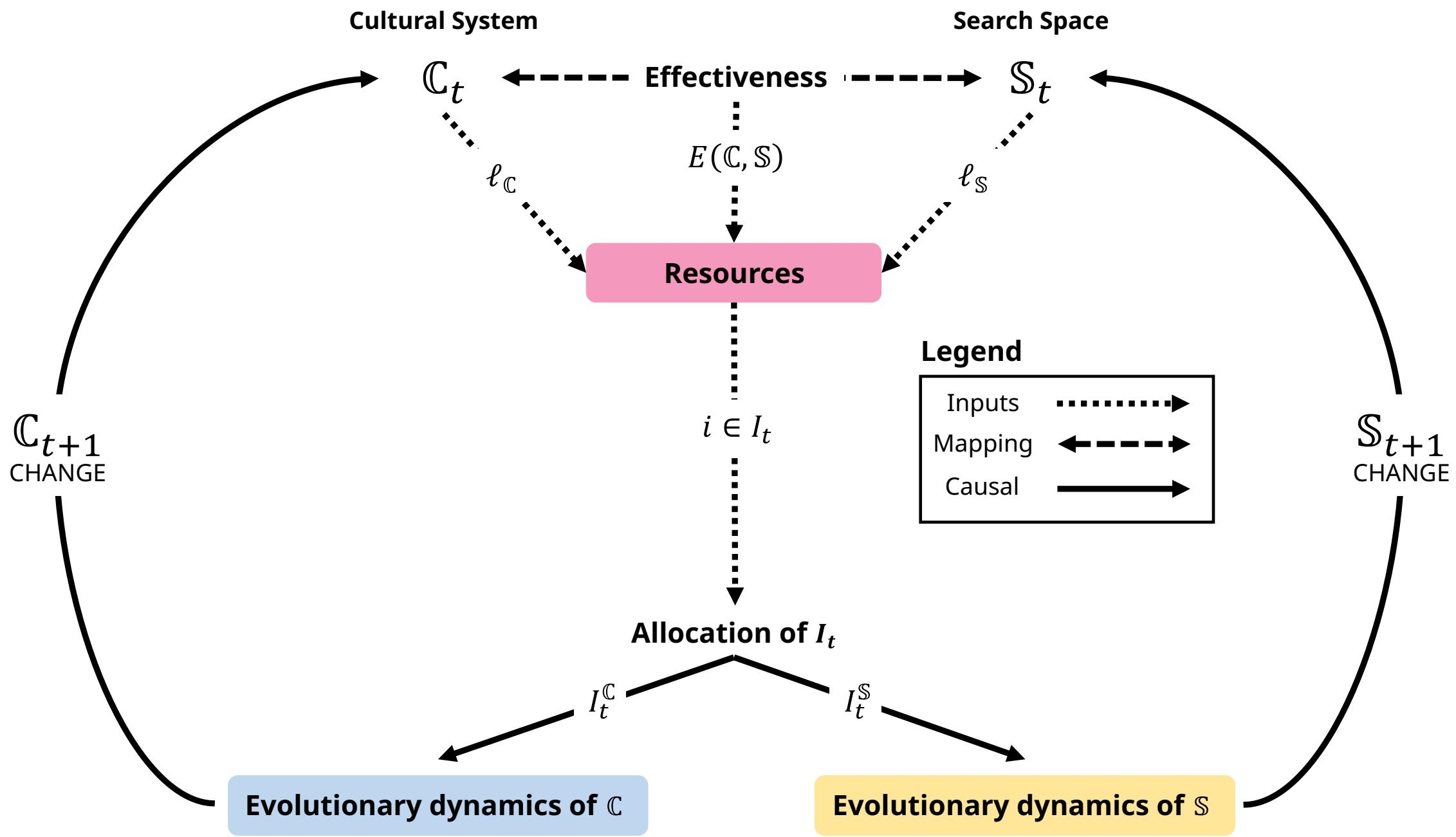

Figure 1: A diagram outlining the general dynamics of bitw0r1d at a single timestep ($t$). For a given $t$, a society measures the effectiveness between $\mathbb{C}_t$ and $\mathbb{S}_t$ (see Section 2.1.1) and their levels of complexity (see Section 2.1.2). The effectiveness of this mapping as well as the complexity of cultural systems ($\ell_{\mathbb{C}}$) and search spaces ($\ell_{\mathbb{S}}$) determine the amount of resources ($R_t$) (see Section 2.2.1). These resources, in form of resource iterations ($i \in I_t$), are then allocated to two processes: one which changes the cultural systems of a society and another which changes the underlying search space (see Section 2.2.2). Both processes follow a cultural evolutionary dynamic of first generating and then adopting changes (see Section 2.1). Our model varies the selection pressures on $\mathbb{C}$ and $\mathbb{S}$ by manipulating the extent to which the adoption of changes are stochastic or deterministic: $\eta$ controls the selection pressures on $\mathbb{C}$ and $\lambda$ controls the selection pressures on $\mathbb{S}$.

Cultural evolutionary dynamics unfold as a discrete-time process ($t \in T$). At a given timestep ($t$), the effectiveness and complexity of $\mathbb{C}$ and $\mathbb{S}$ determine the amount of resources available to a society. These resources are then allocated, in form of resource iterations ($i \in I_t$), to the evolutionary dynamics of generating and adopting changes to $\mathbb{C}$ and $\mathbb{S}$. We allow for three types of change in bitw0r1d: simplifications (which reduce complexity), modifications (which maintain complexity) and expansions (which increase complexity). Whether or not a change is adopted depends on the balance between stochastic and deterministic factors. We manipulate this balance via two parameters: $\eta$ controls the selection pressures on $\mathbb{C}$ and $\lambda$ controls the selection pressures on $\mathbb{S}$.

## 2.1 State variables: Cultural systems ($\mathbb{C}$) and search spaces ($\mathbb{S}$)

At the start of a simulation, a society is initialised with a cultural system ($\mathbb{C}$) and a search space ($\mathbb{S}$). Both $\mathbb{C}$ and $\mathbb{S}$ are operationalised as bitstrings of $N$-length (see [19,20,31,57,89] for similar approaches). Search spaces represent the ensemble of needs, problems and goals facing a society and determine what constitutes an effective cultural system, whereas cultural systems represent the totality of techniques, artifacts, norms and knowledge used to exploit these spaces.

Using bitstrings allows us to represent $\mathbb{C}$ and $\mathbb{S}$ as structured entities that interact with one another and change over time. Structure, in this sense, refers to the distribution and organisation of 0s and 1s in a string (e.g., 11, 01, 111 and 0101 all have different structures). By modelling $\mathbb{C}$ and $\mathbb{S}$ as structured, we capture three important properties. First, the structure of $\mathbb{C}$ or $\mathbb{S}$ at one point in time constrains the possible single-edit changes available to a society (i.e., it restricts the *adjacent possibilities* [90]). Second, the mapping between $\mathbb{C}$ and $\mathbb{S}$ can be measured as a metric distance, which approximates the level of *effectiveness* (i.e., the degree to which cultural systems and search spaces are adapted to one another, see Section 2.1.1). Finally, the string length of $\mathbb{C}$ and $\mathbb{S}$ serves as a proxy for *complexity* (i.e., longer strings denote more complex cultural systems and search spaces, while shorter strings correspond to simpler ones, see Section 2.1.2).

### 2.1.1 Effectiveness: $E(\mathbb{C}, \mathbb{S})$

Effectiveness refers to the mapping between cultural systems and search spaces. Motivating this macro-level focus is the idea that the utility of a cultural trait is derived from its contribution as part of a wider cultural system in addressing the relevant needs, problems and goals of a society. A car, for instance, is only useful insomuch as it fulfils general needs related to transportation, solves specific problems associated with safety and is supported by a system of roads, mechanics and other related infrastructure. This demand is greatly diminished if there is a lack of infrastructure (e.g., no road or access to fuel), minimal safety (e.g., engines or tires susceptible to exploding) and no need for long distance transportation (e.g., living on a small island).

Formally, effectiveness is measured using an inverted form of the normalised Levenshtein distance [91]:

$$E(\mathbb{C}, \mathbb{S}) = 1 - \left( \frac{\mathrm{LD}[\mathbb{C}, \mathbb{S}]}{\max(\ell_{\mathbb{C}}, \ell_{\mathbb{S}})} \right) \tag{2.1}$$

where $\mathrm{LD}[\mathbb{C}, \mathbb{S}]$ measures the minimum number of single-edit operations (insertions, deletions and substitutions) required to transform one string into another:

$$\mathrm{LD}[\mathbb{C}, \mathbb{S}] = \min_{k} \left\{ \sum_{i=1}^{k} c(\Theta_i) \right\} \tag{2.2}$$

Here, $\Theta_i$ denotes the set of all single-edit operations for transforming a string, $k$ represents the number of edit operations in a particular sequence that transforms $\mathbb{C}$ into $\mathbb{S}$ (where $\min_k$ aims to find the sequences with the fewest operations), and $c$ is the unit cost of each operation, where $c(\Theta_i) = 1$. As $\mathbb{C}$ and $\mathbb{S}$ can differ in lengths, we normalise the Levenshtein distance by the longest string, i.e., $\max(\ell_{\mathbb{C}}, \ell_{\mathbb{S}})$, and then invert the values so that $E(\mathbb{C}, \mathbb{S}) \to 1.0$ represents increasingly effective outcomes while $E(\mathbb{C}, \mathbb{S}) \to 0.0$ represents increasingly less effective ones. An optimal outcome is achieved when the structure of a cultural system is equivalent to the structure of a search space, $E(\mathbb{C}, \mathbb{S}) = 1.0$.

Formulating effectiveness as a granular measure allows us to capture the degree to which $\mathbb{C}$ and $\mathbb{S}$ are adapted to one another. This means different mappings of $\mathbb{C}$ and $\mathbb{S}$ can be more or less distant. For example, neither $\mathbb{C}_a = 0100$ nor $\mathbb{C}_b = 0000$ constitute an optimal mapping for $\mathbb{S}_i = 1010$, but it also holds that $\mathbb{C}_a$ is closer to optimal than $\mathbb{C}_b$. Importantly, our measure of effectiveness is agnostic as to the process of change (i.e., whether differences in effectiveness comes from changes to $\mathbb{C}$ or $\mathbb{S}$) as well as the type of change (differences in effectiveness could result from simplifications, modifications or expansions to $\mathbb{C}$ or $\mathbb{S}$). As such, we avoid assuming that (i) improvements in effectiveness solely stem from changes to the cultural system, and (ii) that improvements are concomitant with increases in diversity or complexity.

#### 2.1.2 Complexity: $\ell_{\mathbb{C}}$ and $\ell_{\mathbb{S}}$

Complexity is measured as the string length of cultural systems ($\ell_{\mathbb{C}}$) and search spaces ($\ell_{\mathbb{S}}$). Restricting our definition of complexity to string length allows us to formulate it in terms of computational principles, i.e., the time required (on average) to find a maximally effective outcome given an initial state and an underlying process. Within the general constraints of our model, longer strings can be viewed as more complex than shorter ones because there are a greater number of possible configurations. On average, if we assume $\mathbb{C}$ and $\mathbb{S}$ are maximally distant from one another, i.e., $E(\mathbb{C}, \mathbb{S}) = 0.0$, then a $\ell_{\mathbb{C}} = \ell_{\mathbb{S}} = 20$ will require far fewer changes to find an optimal outcome than $\ell_{\mathbb{C}} = \ell_{\mathbb{S}} = 200$ (see supplementary for a fuller treatment of this point). All else being equal, this assumes that greater computational demands are linked to increases in different types of complexity (see [92] for a thoughtful discussion on this point). Motivating our assumption here is the idea that changes to more complex cultural systems and search spaces should require more computations than simpler ones.

For cultural systems, differences in complexity therefore approximate both the combinatorial sophistication of individual cultural traits as well as the diversity of traits available to a society. A modern smartphone is complex in that it made up of modular, hierarchically organised components (e.g., integrated circuitry, battery technology and software operating systems), and its construction is supported by a variety of manufacturing and distribution processes (e.g., rare earth mineral extraction, semiconductor fabrication, precision assembly and global logistics networks). Meanwhile, differences in the complexity of search spaces not only denote the diversity of needs, problems and goals facing a society, but also the degree to which these interact with one another and are well-defined [93]. Smartphones fulfil diverse needs related to communication, navigation and entertainment in ways that require solving several well-formulated problems and goals (e.g., tradeoffs in processing power versus battery life, the relationship between screen size and pixel density, and thermal management within constrained form factors).

## 2.2 Resources

Evolutionary dynamics in our model are both constrained by and responsible for the production of resources. Differences in resources translate into differences in the ability of societies to change $\mathbb{C}$ and $\mathbb{S}$. Gains in resources therefore aim to approximate increases in population size [44], new technologies that enhance a society's capacity to store and process information [33], or improvements in the harnessing and utilising of new energy sources [35]. We conceptualise this as a two-step process of first producing (Section 2.2.1) and then allocating resources (Section 2.2.2).

### 2.2.1 Producing resources: $R(\mathbb{C}, \mathbb{S}, E)$

The production of resources ($R$) is governed by the interaction between $\ell_{\mathbb{C}}$, $\ell_{\mathbb{S}}$ and $E$ (which denotes a simplified notation for effectiveness):

$$R(\ell_{\mathbb{C}}, \ell_{\mathbb{S}}, E) = \underbrace{\ell_{\mathbb{S}} E}_{\text{benefits}} - \underbrace{\ell_{\mathbb{C}}(1 - E)}_{\text{costs}} \tag{2.3}$$

where $\ell_{\mathbb{S}} E$ represents the benefits of effectively exploiting a search space and $\ell_{\mathbb{C}}(1 - E)$ captures any costs arising from ineffective cultural systems.

We make two main assumptions here. First, gains in resources are bounded by the complexity of the search space ($\ell_{\mathbb{S}}$), with $E(\mathbb{C}, \mathbb{S})$ determining how close a society is to realising this resource potential. As such, more complex search spaces are associated with a greater resource potential than simpler ones (i.e., the possible ways in which needs, problems and goals can be converted into resources). Second, the costs for ineffectiveness, $1 - E(\mathbb{C}, \mathbb{S})$, scale with the complexity of cultural systems ($\ell_{\mathbb{C}}$). As cultural systems grow in complexity, a society will incur increased costs if it is maintaining techniques, norms, artifacts and knowledge that do not effectively contribute to the production of resources (i.e., capabilities that use more resources than they produce).

### 2.2.2 Allocating resources: $I_t^{\mathbb{C}}$ and $I_t^{\mathbb{S}}$

At each timestep, a society allocates a set of resource iterations ($i \in I_t$) for changing its cultural system ($I_t^{\mathbb{C}}$) or its search space ($I_t^{\mathbb{S}}$). Each resource iteration denotes an indexed position, $I_t = \{i_0, ..., i_n\}$, where the total number of iterations available is bounded by the resources produced at the previous timestep, i.e., $\lfloor R_{t-1} \rceil$. As a simplifying assumption, resource allocation to $\mathbb{C}$ or $\mathbb{S}$ remains unbiased, i.e., $P_{\text{allocate}} \sim \text{Bernoulli}(0.5)$, meaning there is no inherent preference for making changes to $\mathbb{C}$ or $\mathbb{S}$.

One consequence of our approach is that changes to $\mathbb{C}$ and $\mathbb{S}$ are interleaved with one another. Imagine, for instance, there are six resource iterations available to a society at $t$: resource iterations 1, 2 and 5 could be allocated to changing $\mathbb{C}$ (i.e., $I_t^{\mathbb{C}} = \{1, 2, 5\}$) whereas the remaining iterations are allocated to changing $\mathbb{S}$ (i.e., $I_t^{\mathbb{S}} = \{3, 4, 6\}$). As such, a change to $\mathbb{C}$ at resource iteration 5 is made in reference to the state of $\mathbb{S}$ at iteration 4, while a change to $\mathbb{S}$ at resource iteration 6 is made in reference to the state of $\mathbb{C}$ at iteration 5.

Arguably, by assuming the order of changes are interleaved, our model better reflects the fact that changes in cultural systems often lead to changes in the search space, and vice versa. For instance, we commonly observe instances where the adoption of a cultural solution triggers a cascade of new problems to solve (e.g., a society dependent on fire for purposes of warmth and protection needs to solve additional problems associated with finding fuel, smoke ventilation and fire containment), and the adoption of a goal serves as a catalyst for novel solutions (e.g., the goal of actively increasing farming yields has led to the creation of synthetic fertilisers, pesticides and advanced irrigation systems).

### 2.3 Cultural evolutionary dynamics

Cultural evolutionary dynamics are modelled as two discrete-time processes: one which changes a cultural system $(\mathbb{C}_0, ..., \mathbb{C}_t)$ and another which changes a search space $(\mathbb{S}_0, ..., \mathbb{S}_t)$. For each timestep, the resources allocated to cultural systems $(I_t^{\mathbb{C}})$ and search spaces $(I_t^{\mathbb{S}})$ determine the number as well as the order of changes available to the dynamics (see Section 2.2.2). We can describe the accumulated changes to $\mathbb{C}$ and $\mathbb{S}$ at $t+1$ in terms of two general functions:

$$\begin{cases} \mathbb{C}_{t+1} = f(\mathbb{C}_t, \mathbb{S}_i, \eta, I_t^{\mathbb{C}}) = \mathbb{C}_t + \sum_{i \in I_t^{\mathbb{C}}} \Delta\mathbb{C}_i \\ \mathbb{S}_{t+1} = g(\mathbb{C}_i, \mathbb{S}_t, \lambda, I_t^{\mathbb{S}}) = \mathbb{S}_t + \sum_{i \in I_t^{\mathbb{S}}} \Delta\mathbb{S}_i \end{cases} \tag{2.4}$$

where $\mathbb{C}_t$ and $\mathbb{S}_t$ are the states of the cultural system and the search space at time $t$, $\mathbb{C}_i$ and $\mathbb{S}_i$ are the states of the cultural system and the search space at resource iteration $i$, $I_t^{\mathbb{C}}$ and $I_t^{\mathbb{S}}$ are the allocated resource iterations available for changing $\mathbb{C}_t$ and $\mathbb{S}_t$, and $\eta$ and $\lambda$ are the parameters for controlling the strength of selection on $\mathbb{C}$ and $\mathbb{S}$.

Equation 2.4 tells us that the accumulated changes to cultural systems and search spaces are resource constrained: the number of possible changes at $t+1$ is proportional to the allocated resource iterations of $I_t^{\mathbb{C}}$ and $I_t^{\mathbb{S}}$. The selection parameters $\eta$ and $\lambda$ determine which changes are adopted within those constraints. As changes are indexed by $i$, selection dynamics for cultural systems are made in reference to the state of the search space at that resource iteration (i.e., $\mathbb{S}_i$), and vice versa for selection dynamics in the context of search spaces (where a selected change is made in reference to $\mathbb{C}_i$). Here, a single resource iteration corresponds to an evolutionary process of first generating (Section 2.3.1) and then adopting changes (Section 2.3.2) to either $\mathbb{C}$ or $\mathbb{S}$.

**2.3.1 Generative dynamics: $\theta_j$**

Generating a change is possible via three general operators: to remove a randomly chosen bit (e.g., *remove*: 1011 → 101), to flip a randomly chosen bit to its Boolean complement (e.g, *modify*: 1011 → 1111) or to introduce a new randomly chosen bit at a randomly assigned position (e.g., *expand*: 1011 → 10111). The choice of operation always remains unbiased in the model, i.e., the probability of choosing an operation is $\theta_j \in \{\theta_{\text{simplify}}, \theta_{\text{modify}}, \theta_{\text{expand}}\}$ with $P(\theta_j) = \frac{1}{3}$.

By modelling the generation of changes as unbiased, we can investigate the role played by stochasticity in driving long-term evolutionary dynamics. This means any deterministic factors are found in the adoption of changes to $\mathbb{C}$ or $\mathbb{S}$ (see Section 2.3.2). It is worth noting that our assumption here only pertains to population-level factors. Certain individuals within a society could be biased towards introducing specific types of change, but we assume these biasing factors are non-systematic at the population-level.

**2.3.2 Adoption dynamics: $\eta$ and $\lambda$**

Once a change is introduced, its adoption is contingent on whether the dynamics are stochastic or deterministic. For cultural systems, the adoption of a change at a given resource iteration is underpinned by the following:

$$P(\mathbb{C}'|\mathbb{C}_i, \mathbb{S}_i, \eta) = \underbrace{(1-\eta)}_{\text{stochastic}} + \underbrace{\eta \cdot 1[E(\mathbb{C}', \mathbb{S}_i) > E(\mathbb{C}_i, \mathbb{S}_i)]}_{\text{deterministic}} \tag{2.5}$$

where $\mathbb{C}_i$ is the current state of a cultural system at resource iteration $i$ for timestep $t$, $\mathbb{C}'$ represents a proposed change to this system at $i$, and $\mathbb{S}_i$ is the current state of the search space at $i$. $\eta$ is a parameter $\in [0, 1]$ that manipulates the probability a change is adopted stochastically or deterministically.

If $\eta = 0.0$, then the evolution of a system is purely stochastic and reduces to $P(\mathbb{C}'|\mathbb{C}_i, \mathbb{S}_i, \eta) = (1-\eta) = 1.0$, i.e., a change is adopted irrespective of whether or not it increases effectiveness. Conversely, for $\eta = 1.0$, the evolutionary dynamics are purely deterministic and the outcome depends on the indicator function $1[\cdot]$: here, if $\mathbb{C}'$ is more effective for exploiting a search space, as denoted by $E(\mathbb{C}', \mathbb{S}_i) > E(\mathbb{C}_i, \mathbb{S}_i)$, then $\mathbb{C}'$ is adopted as the new state of the system at the next resource iteration ($\mathbb{C}_i = \mathbb{C}'$). Otherwise, if $E(\mathbb{C}', \mathbb{S}_i) \leq E(\mathbb{C}_i, \mathbb{S}_i)$ a society remains with the existing state of $\mathbb{C}_i$. Intermediate values, $\eta = (0.0, 1.0)$, incorporate some mixture of deterministic and stochastic factors into the dynamics of adopting changes. As such, the structure of cultural systems will to some extent reflect both randomly adopted changes as well as changes selected on the basis of improving effectiveness. Modelling the dynamics in this way assumes the adoption of changes is endogenous and incremental. A cultural

system adapts insomuch as the adoption of changes helps address the needs, problems and goals of a society.

Parallel dynamics hold for changes to a search space:

$$P(\mathbb{S}'|\mathbb{C}_i, \mathbb{S}_i, \lambda) = \underbrace{(1-\lambda)}_{\text{stochastic}} + \underbrace{\lambda \cdot 1[E(\mathbb{C}_i, \mathbb{S}') > E(\mathbb{C}_i, \mathbb{S}_i)]}_{\text{deterministic}} \tag{2.6}$$

except $\lambda$ now controls the stochastic and deterministic forces acting upon the adoption of changes to a search space ($\mathbb{S}_i$). Values of $\lambda \to 0.0$ increasingly adopt random changes to the search space, whereas values of $\lambda \to 1.0$ increasingly evaluate and adopt changes to $\mathbb{S}_i$ that improve effectiveness. This means that search spaces evolve by changing the needs, problems and goals of a society. Some of these changes are random and others are selected when $\lambda \in (0.0, 1.0]$. A search space culturally adapts by adopting changes that improve effectiveness, i.e., the needs, problems and goals are restructured to more effectively map onto the existing cultural capabilities of a society.

We can think of different parameter values for $\eta \in [0, 1]$ and $\lambda \in [0, 1]$ as determining the extent to which cultural systems and search spaces are coupled to one another and capable of co-evolution. Selecting improvements for effectiveness is therefore possible by changing either $\mathbb{C}$ or $\mathbb{S}$. This represents a process by which cultural systems adapt to search spaces, and vice versa. For example, a society that wants to land a person on the Moon can adapt their cultural systems to utilise rockets for escaping the atmosphere, computers for rapidly performing complex trajectories and life support systems for surviving in the vacuum of space. All of these represent finely-tuned solutions to a variety of needs, problems and goals related to space travel. However, societies are not just passively responding to some exogenous set of demands. Instead, individuals and groups play an active role in constructing their search spaces: through institutions, markets and other forms of collective deliberation, societies select the needs they prioritise, the problems that require solving, and the goals deemed worthy of pursuing. Indeed, the goal of landing a person on the moon before the end of the 1960s was formulated in the context of existing technological (e.g., the availability of rocketry), scientific (e.g., knowledge of orbital mechanics) and economic (e.g., the use of supply chains) capabilities [94].

## 2.4 Simulation runs and boundary conditions

For the reported simulation runs, societies are initialised with randomly sampled cultural systems and search spaces of $\ell_{\mathbb{C}} = \ell_{\mathbb{S}} = 2$ (minimum level of complexity is $\ell = 1$). Parameter combinations of $\eta \in (0, 1)$ and $\lambda \in (0, 1)$ consist of $K_{\text{sim}} = 1000$ simulations. This assumes there is always some degree of co-evolution between $\mathbb{C}$ and $\mathbb{S}$ and allows us to investigate a range of scenarios: from strongly stochastic ($\eta = \lambda = 0.01$) to strongly deterministic ($\eta = \lambda = 0.99$). A simulation halts under one of three conditions: (i) the maximum number of timesteps is reached ($\max_t = 10,000$), (ii) the upper-bound of cultural complexity is either matched or exceeded ($\max_{\ell_{\mathbb{C}}} \geq 10,000$), or (iii) societies hit an absorbing barrier having exhausted their resources ($R_t \leq 0$).

For (iii), absorbing barriers refer to a point of no return from which a system cannot recover [88]. In the context of our model, this approximates a situation where cultural evolutionary dynamics are unsustainable for producing resources. Importantly, we assume absorbing barriers only come into effect after societies have exhausted an initial resource endowment ($R_{\text{endow}} = 100$). Introducing an endowment provides a warm up period for the dynamics and mitigates the initial conditions from too strongly influencing the outcomes (see supplementary for different initial endowments).

Societies only draw upon this endowment when the resources produced at the previous timestep are $\lfloor R_{t-1} \rceil \leq 0$. Here, a society is allocated a single resource so that $I_t = 1$, and the endowment is reduced by the corresponding loss in resources (e.g., a loss of $\lfloor R_{t-1} \rceil = -10$ would mean the initial endowment is updated to $R_{\text{endow}} = 100 - 10 = 90$). A special case exists where $\lfloor R_{t-1} \rceil = 0$: we treat this as costly and subtract $-1$ from the endowment. Once an endowment is exhausted, $R_{\text{endow}} \leq 0$, a society is no longer buffered against absorbing barriers and the dynamics terminate at the next timestep (i.e., $I_t = \emptyset$).

### 2.5 Code availability

All code, data and supplementary material for reported runs are available from the following repository: https://github.com/j-winters/bitw0r1d.

## 3 Results

A general finding of our model is that absorbing barriers represent the majority outcome across all parameter configurations (see Figure 2). Without resources, a society cannot make changes to either its cultural system or search space. In total, $\approx 98\%$ of simulation runs terminated as a result of an absorbing barrier, suggesting this constitutes a hard constraint on the emergence of open-ended growth. Of these, runs that most successfully avoided absorbing barriers corresponded to $\eta = 0.95$ and $\lambda = 0.80$ ($\approx 11.4\%$ reaching $10,000$ generations) while the runs with the highest average survival rate are found at $\eta = 0.99$ and $\lambda = 0.99$ (reaching an average of $\approx 3060$ timesteps).

The likelihood of hitting an absorbing barrier increases as the dynamics become increasingly stochastic. At one extreme of the parameter space ($\eta = \lambda = 0.01$), all simulation runs terminate as a result of absorbing barriers. Similar findings hold even when lowering the degree of stochasticity (e.g., $\eta = 0.2$ and $\lambda = 0.6$). In these cases, excessive stochasticity causes $\mathbb{C}$ and $\mathbb{S}$ to evolve relatively independently of one another, with changes largely following a random walk (i.e., cultural drift). This quickly results in collapse as societies are unable to discover effective outcomes and produce a net gain in resources.

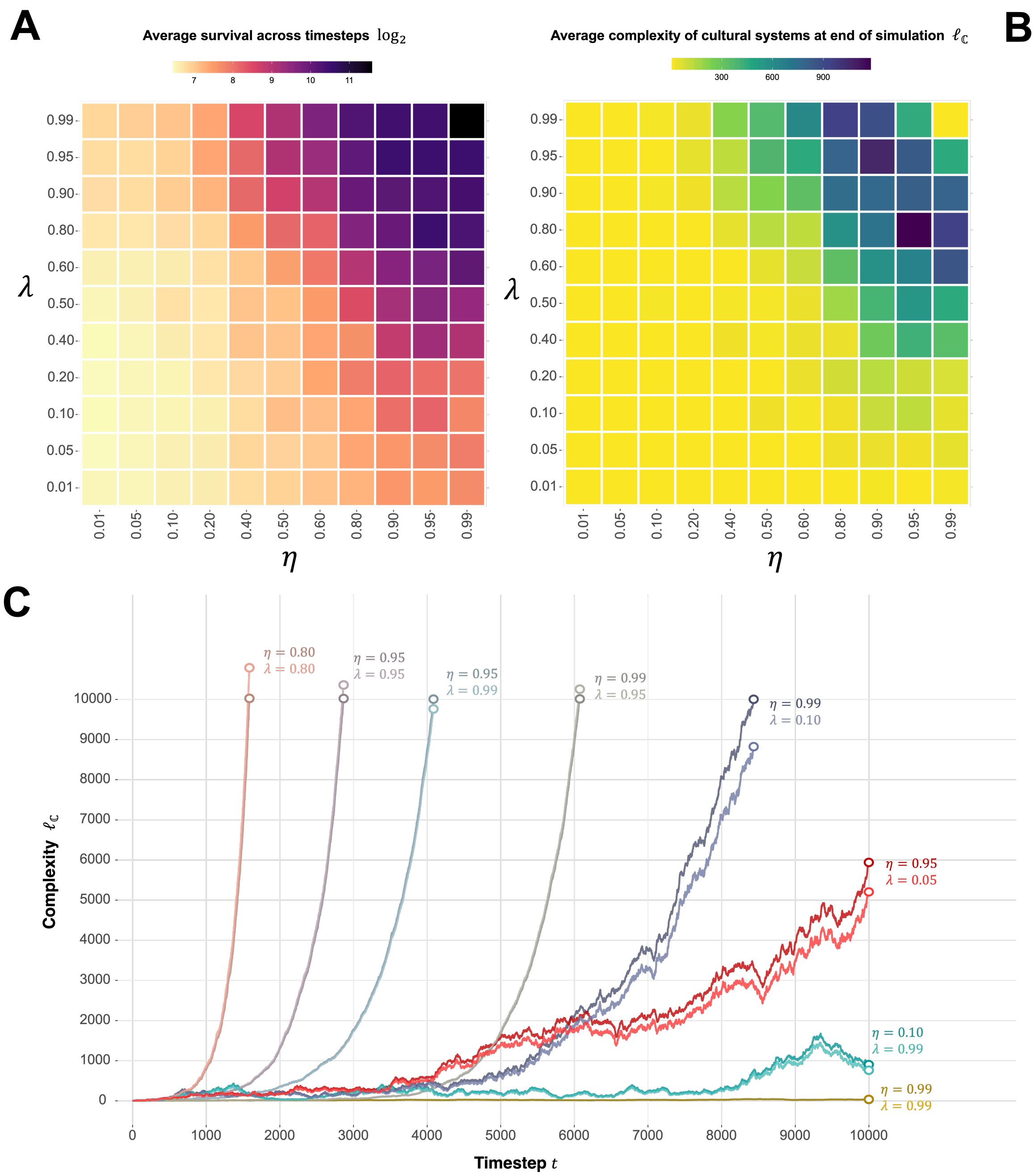

Figure 2: **A**: Heatmap showing the average survival time ($\log_2$) for all simulation runs across $\eta$ and $\lambda$ (note that runs which matched or exceeded $\ell_{\mathbb{C}} = 10,000$ were treated as reaching the full 10,000 timesteps. **B**: Heatmap showing the average complexity of cultural systems ($\ell_{\mathbb{C}}$) for all simulation runs across $\eta$ and $\lambda$ at the point of termination. **C**: Selected simulation runs showing a range of dynamics. Coloured lines represent the complexity of cultural systems (denoted with $\eta$) and search spaces (denoted with $\lambda$) for different parameter values.

At the other extreme ($\eta = \lambda = 0.99$), stronger selection pressures are more capable of sustaining co-evolutionary dynamics, but many simulation runs remain trapped in low complexity states for $\mathbb{C}$ and $\mathbb{S}$. This is due to the co-evolutionary dynamics maintaining a tight coupling where the adoption of changes in $\mathbb{C}$ are made in reference to improving effectiveness for $\mathbb{S}$, and vice versa. We can think of this in terms of emergent local optima that confine the dynamics to simpler cultural systems and search spaces. For the majority of runs, the dynamics will either remain in these simple states or hit an absorbing barrier. However, as Figure 3 illustrates, a minority do manage to escape and undergo open-ended growth.

Avoiding absorbing barriers is also possible when one process is strongly deterministic ($\eta = 0.99$) and the other is strongly stochastic ($\lambda = 0.01$). This creates a moving target where the deterministic adoption of changes in one process tracks the stochastic changes of the other. Simulation runs in these instances are capable of reaching $10,000$ timesteps, but the stochastic nature of the dynamics means we cannot definitively state for a given simulation run whether $\mathbb{C}$ and $\mathbb{S}$ will keep co-evolving indefinitely or eventually hit an absorbing barrier (see Figure 3). Long-term sustainability in such circumstances somewhat depends on the complexity of $\mathbb{C}$ and $\mathbb{S}$. Reaching more complex states is advantageous insomuch as a single adverse change represents a proportionally smaller reduction to the overall effectiveness. Crucially, this robustness-enhancing effect of complexity only holds when deterministic factors are present and strong enough to act as a countervailing force: here, adopting changes that improve effectiveness counteract stochastic perturbations, preventing drift-like dynamics from dramatically depleting the resources available to a society.

Notably, important differences exist depending on whether these stronger selection pressures are applied to cultural systems (e.g., $\eta = 0.99; \lambda = 0.01$) or search spaces (e.g., $\eta = 0.01; \lambda = 0.99$). As Figure 3 demonstrates, stronger selection pressures for cultural systems result in high levels of sustainability and more complex outcomes than when similarly powerful selection pressures are applied to search spaces. This asymmetry is largely driven by our formulation of the resource function. While strong selection pressures on $\mathbb{C}$ help mitigate any excess costs by tracking changes to $\mathbb{S}$, weak selection pressures on $\mathbb{C}$ are especially costly when a random change that lowers effectiveness also increases cultural complexity ($\ell_{\mathbb{C}}$).

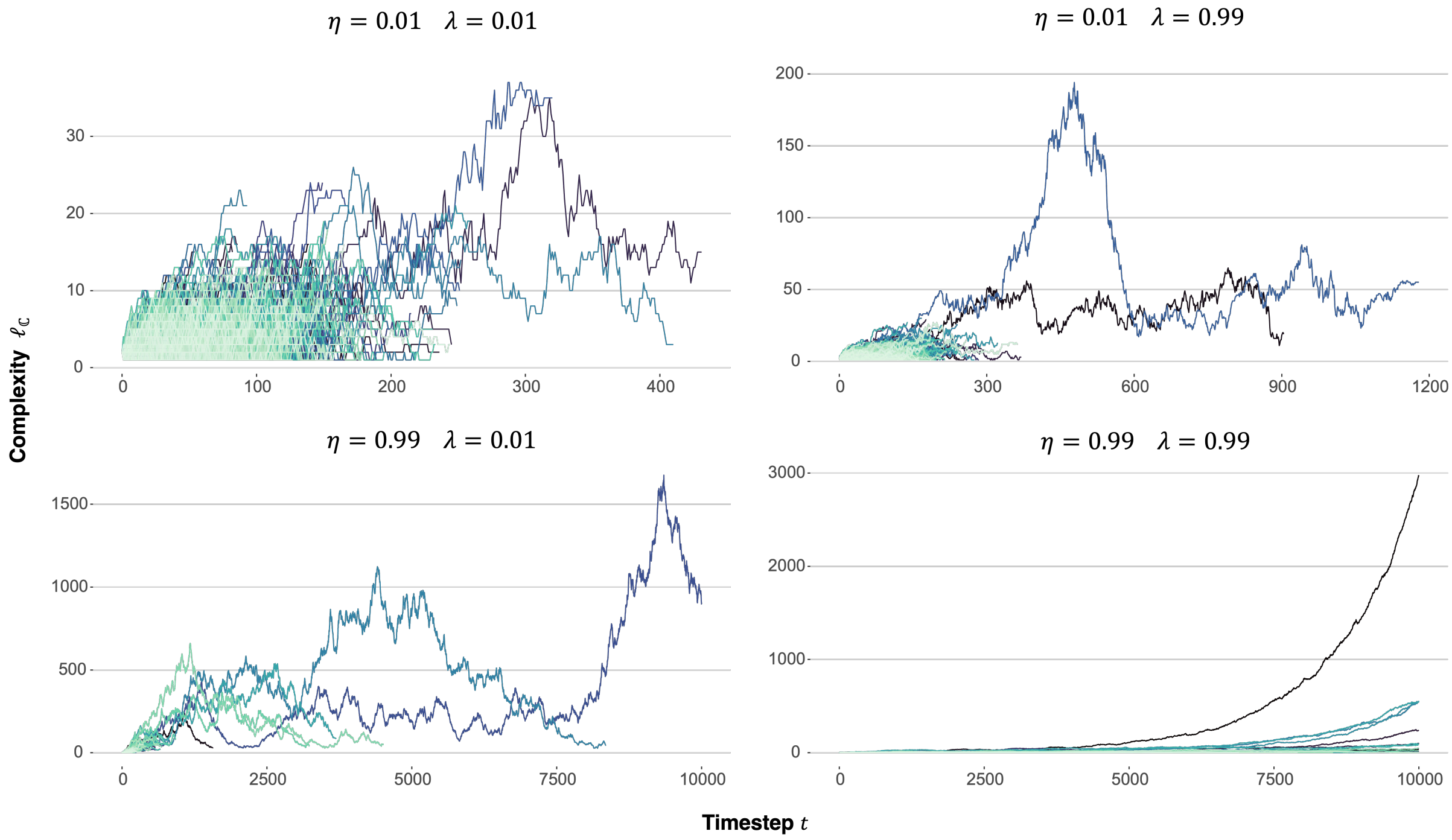

Figure 3: All simulation runs (coloured lines) showing the complexity for cultural systems ($\ell_{\mathbb{C}}$) for $\eta \in [0.01, 0.99]$ and $\lambda \in [0.01, 0.99]$. Note that the axes vary between facets of different parameter values. When $\eta = 0.01$, none of the runs manage to survive for the full $10,000$ timesteps and are restricted to relatively simple cultural systems. By contrast, when $\eta = 0.99$, a minority of runs manage to reach $t = 10,000$, with $\eta = 0.99$ and $\lambda = 0.99$ exhibiting open-ended growth.

A critical threshold exists in our model where the dynamics transition to an open-ended regime, i.e., the dynamics avoid absorbing barriers and undergo long-term growth in the complexity of cultural systems and search spaces. At this point, simulation runs either reach $t = 10,000$ timesteps or exceed the upper-limit of the simulation ($\ell_{\mathbb{C}} = 10,000$). Different parameter combinations of $\eta$ and $\lambda$ influence the timing, frequency and rate of this growth (see Figure 2 C and top row of Figure 4). For some runs, open-ended evolution is late emerging, extremely rare and slow growing (e.g., Figure 2 C for $\eta = 0.95$ and $\lambda = 0.05$) whereas other runs are early emerging, more common and undergo accelerated growth (e.g., Figure 2 C for $\eta = 0.80$ and $\lambda = 0.80$). Minimally, for simulation runs across all relevant parameter values, the emergence of open-ended evolution depends on three general conditions.

First, the dynamics exist in a far-from-equilibrium state, resulting from a balance of stochastic and deterministic factors. Specifically, the parameter values most conducive to open-endedness are those where stochastic factors play a prominent role in shaping both $\mathbb{C}$ and $\mathbb{S}$, but are generally weaker than deterministic ones, e.g., compare $\eta = 0.40$ and $\lambda = 0.40$ with $\eta = 0.90$ and $\lambda = 0.90$ (see top row of Figure 4). Having some degree of stochasticity prevents the dynamics from settling into a stable equilibrium, while the presence of strong selection pressures helps maintain effectiveness in the face of these stochastic perturbations (see bottom row of Figure 4).

Second, in this far-from-equilibrium state, selection is strong enough to preferentially adopt expansionary changes. Although the generation of changes is equiprobable, with $P(\theta_j) = \frac{1}{3}$, there are many more ways to expand $\mathbb{C}$ or $\mathbb{S}$. A greater diversity of possible states means that on average there is a greater chance of discovering an effective outcome via expansion than via modification or simplification. As a result, we should expect a net growth in complexity when (i) stochastic factors continually perturb the dynamics and open up the possibility for improving effectiveness (see previous paragraph), and (ii) there are sufficient resources to discover and adopt expansionary changes (see next paragraph).

Third, discovering these expansionary changes requires a society increases its production of resources. In low complexity states, the advantages of expansion can remain masked due resource scarcity, with fewer resources translating into fewer evolutionary opportunities for changing $\mathbb{C}$ or $\mathbb{S}$. As a consequence, the likelihood of discovering a beneficial expansion is constrained by the probability of generating this type of change in the first place. However, as the dynamics transition to more complex states for $\mathbb{C}$ and $\mathbb{S}$, gains in resources increase the number of iterations, which, in turn, amplifies the chances of adopting expansionary changes. Under the right conditions, this causes open-ended growth to accelerate and exceed the upper-limit of our simulation (i.e., $\ell_{\mathbb{C}} \geq 10,000$).

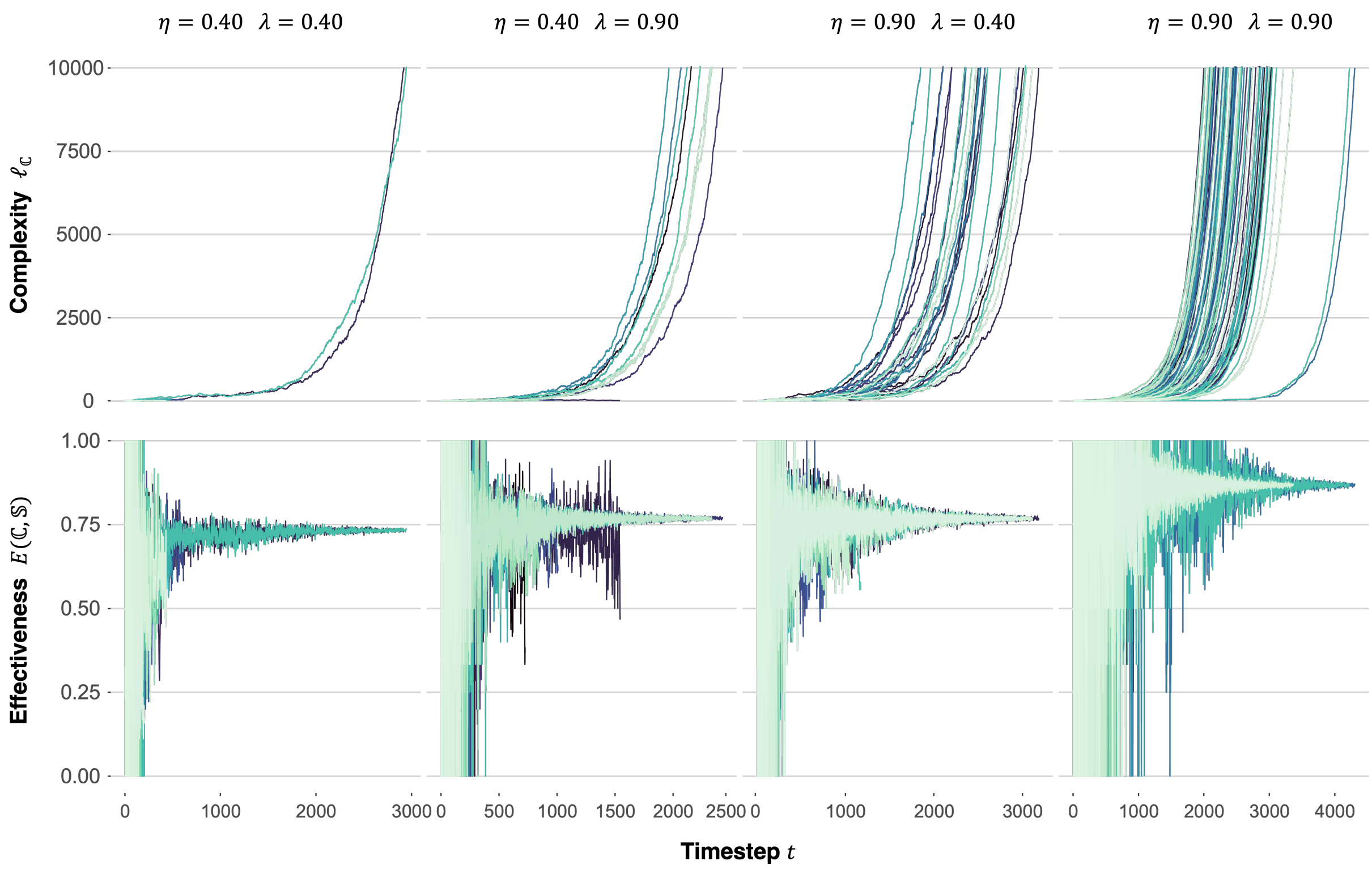

Figure 4: **Top row:** All simulation runs (coloured lines) showing the complexity of cultural systems ($\ell_{\mathbb{C}}$) for $\eta \in [0.4, 0.9]$ and $\lambda \in [0.4, 0.9]$. **Bottom row:** The same simulation runs, but showing the level of effectiveness reached, $E(\mathbb{C}, \mathbb{S})$.

## 4 Discussion

Our paper started from a relatively simple premise: that *cultural systems* ($\mathbb{C}$, the techniques, artifacts, norms and knowledge available to a population) and *search spaces* ($\mathbb{S}$, the ensemble of needs, problems and goals facing a society) co-evolve with one another. Unlike previous macro-level models, which focus on explaining the evolution of cultural systems and abstract away from explicitly representing search spaces, we modelled cultural systems and search spaces as evolving entities that are shaped by a mixture of stochastic and deterministic factors. In our model, adopting changes not only impacts the mutual fit between cultural systems and search spaces, it also leads to changes in structure that expand, maintain or reduce complexity. By manipulating the extent to which the adoption of these changes is stochastic or deterministic, we investigated the conditions in which co-evolutionary dynamics are sustainable and lead to the emergence of open-ended evolution.

Modelling $\mathbb{C}$ and $\mathbb{S}$ as co-evolving forces us to think seriously about processes that are typically idealised away and placed in a black box. In bitw0r1d, we assume humans interact with the world by generating and adopting changes to their needs, problems and goals. This allows individuals to evaluate their own behaviours, as well as the behaviours of others, in reference to underlying psychological constructs [73–75]. In aggregate, collections of needs, problems and goals constrain the structure of cultural systems by determining what does and does not constitute an effective change to the cultural traits of a society. Our theoretical aim here is to employ search spaces as a conceptual bridge between problem spaces in cognitive science [74] and adaptive landscapes in evolutionary theory [59]. Although it is relatively uncontroversial to model cultural systems as evolving [8,16,17], we also argue similar dynamics extend to search spaces (see [95] for a recent review of similar debates in the context of *open-ended learning*). Assuming that changes to the search space are evolutionary might appear to contradict the expectation that adaptive pressures are linked to the local environment (e.g., [55]). However, our model remains agnostic about the origin of these adaptive pressures, and is compatible with two general interpretations.

One reading of our model is that environmental constraints determine the range of available needs, problems and goals. Climatic variability [96], migration to different ecologies [97] and niche construction [98] each represent ways in which changes in the environment interact with search spaces. Under this reading, selecting changes to a search space entails discovering which needs, problems and goals are relevant within a wider space of latent possibilities, i.e., those changes that addressable given the current state of the environment and the existing cultural system. For instance, some sources of infertility may be present in populations $50,000$ years ago, existing as an objective biological problem. Yet, recognising that infertility is a problem worth addressing, identifying a relevant set of well-defined sub-problems and then taking steps to develop actionable solutions, is not culturally possible without advanced scientific knowledge, access to biomedical technology and significant economic resources. Another reading is that societies not only select needs, problems and goals through their culture, but actively provide or construct new ones [99]. Architectural problems (e.g., how to build a pyramid), scientific goals (e.g., landing a person on the Moon) and economic needs (e.g., establishing a stable international currency exchange) are all instances where changes to a search spaces are constructed and then adopted in reference to existing cultural and technological capabilities of a society. Of course, these are two extremes, and it is likely that some problems objectively exist in the world, awaiting to be discovered and solved, while others are culturally constructed as a result of intentional activities and unintended consequences [100,101].

A key contribution of this paper is to link the underlying co-evolutionary dynamics to the costs and benefits of producing resources. Two routes exist in our model for increasing the resources available to a society: by improving the effectiveness of cultural systems or through expanding the search space. We show that both routes are only possible under sufficiently strong selection pressures. Although maintaining a certain level of effectiveness is necessary for minimising the costs of supporting a complex cultural system, open-ended growth only emerges when there is a sustained expansion of the search space. Search spaces play a critical role in this sense because they serve as an upper-limit on the maximum resource potential. Conceptually, expansion can be thought of as recognising and adopting needs, problems and goals associated with gains in the total available resources. This is clearly a simplification in that the adoption of new needs, problems and goals is not strictly concomitant with increases in resources. Nevertheless, it is also the case that resources are an ultimate constraint on technological progress [60], which suggests that any long-term expansion has to eventually resolve issues concerning computational and energy demands [34,35,67,80].

The co-evolutionary dynamics in the model are sensitive to the selection pressures acting on both cultural systems ($\eta$) and search spaces ($\lambda$). The strength of these pressures is determined by the relative balance of stochastic and deterministic factors. As such, we can think of $\eta$ and $\lambda$ as a map of where co-evolutionary dynamics are and are not sustainable. At one extreme, in regions dominated by stochasticity (e.g., $\eta = \lambda = 0.01$), an inability to produce resources very quickly leads to an absorbing barrier: a boundary condition on the ability of societies to generate and adopt changes to their cultural systems and search spaces. Our findings here place a very clear limit for explanations that rely on stochasticity as a primary driver of open-endedness [102,103] and contribute to wider debates over the role of neutral models in cultural evolution [104–106]. We show that in a mutualistic co-evolutionary scenario, where survival is contingent on the coupling of co-evolving entities, any factor that causes a decoupling runs the risk of collapse. Stochasticity is a relevant contributing factor, but only when it exists alongside selection pressures that help maintain effectiveness.

Minimally, open-ended growth is only possible so long as stochastic factors keep the co-evolutionary dynamics in a far-from-equilibrium state, and deterministic factors lead to the increased adoption of expansionary changes. Regions of the parameter space most conducive to open-endedness are those characterised by significant selection pressures on cultural systems *and* search spaces. Here, co-evolutionary dynamics play a compensatory role: cultural systems adapt to the structure of search spaces and search spaces adapt to the structure of cultural systems. There are many possible parameter combinations that give rise to open-ended growth in this respect. We suggest that the most plausible route is one in which the selection pressures are gradually increasing in both cultural systems and search spaces. Not only does this account provide a gradualistic pathway to emergence, as the values of $\eta$ and $\lambda$ make it easier to traverse from unsustainable to sustainable co-evolutionary dynamics, it also corresponds to regions of the parameter space where open-ended growth is most likely to undergo progressive acceleration.

Accelerated growth is arguably the defining characteristic of the last few centuries of scientific, technological and economic progress [45,60,107–109]. Growth accelerates in our model when there are sufficient resources to search for and discover beneficial expansions. A subtle difference between our model and others is that we show how the advantages of expansion can remain masked when selection is weak for one process (e.g., $\eta = 0.99$ and $\lambda = 0.01$). Unmasking this advantage, then, requires significant selection pressures in both cultural systems and search spaces (e.g., $\eta = 0.40$ and $\lambda = 0.40$). It is only under these conditions that the co-evolutionary dynamics are powerful enough to overcome initial resource limitations and to amplify the opportunities for generating and adopting expansionary improvements. Analogous constraints exist in cultural evolutionary models where differences in the effective population size impacts the spread of adaptive cultural traits [44,110]. An advantage of our approach is that differences in resources are not limited to demographic factors, but could also vary due to the availability of cognitive technologies for processing information [33,111] (e.g., societies with and without writing systems) as well as the utilisation of energy sources [35,67] (e.g., societies which can and cannot harness the energy potential of coal).

Like all models, which are always wrong and sometimes useful [112], our model relies on several theoretical and methodological simplifications. Some possible critiques here concern the existence of an initial resource endowment, the linear relationship between search space complexity and its resource potential, the assumption that resource allocation is unbiased, and the restricted focus on a single society. First, to help avoid absorbing barriers, we assume societies start out with an initial resource endowment ($R_{\text{endow}} = 100$). Methodologically, our choice of $R_{\text{endow}} = 100$ is a convenient simplification, which avoids the initial conditions from too strongly constraining the dynamics (see the supplementary for different resource endowments). Theoretically, we should expect societies to have access to some initial resources that are potentially non-renewable and susceptible to depletion. For example, we know that during the Pleistocene there was an abundance of megafauna, with technological regressions and advancements linked to the overexploitation of these resources [108,113]. Societies can therefore temporarily exceed their natural carrying capacity by drawing upon excess resources. The challenge, when viewed through the lens of our model, is to either develop sustainable practices before exhaustion or culturally adapt via the co-evolutionary dynamics.

Our second assumption is that the complexity of a search space forms a linear relationship with its resource potential: each additional expansion of $\mathbb{S}$ yields a unit increase in the maximum obtainable resources (e.g., a $\ell_{\mathbb{S}} = 10$ produces a maximum of $R = 10$ whereas a $\ell_{\mathbb{S}} = 100$ produces a maximum of $R = 100$). Relaxing this assumption, to consider a greater diversity of resource functions, would allow us to model situations where search spaces of the same complexity differ in their maximum resource potential. By incorporating such extensions, future work could manipulate the extent to which resource functions form smooth (adjacent search spaces share similar resource functions) or rugged distributions (adjacent search spaces have different resource functions). This represents a tantalising prospect for bridging the tunable topologies of N/K fitness landscapes [37,57] and the open-ended search spaces of bitw0r1d. Navigating these topologies would likely lead to a greater variety of dynamics and could help us understand the situations in which open-ended growth undergoes long-term stagnation [114,115].

A third assumption pertains to the unbiased allocation of these resources. Maintaining an equitable distribution of resources assumes societies always have an equal chance of changing their cultural systems or search spaces. This essentially builds-in a resolution to any explore-exploit dilemmas when allocating resources [116–118]. However, we do not know if this represents an optimal allocation strategy, and neither do we know whether societies are capable of reaching this balanced allocation in the first place. A more sophisticated model could treat societies as Bayesian learners [119] that dynamically update how resources are allocated. Different priors would represent different initial resource allocation strategies, i.e., hypotheses over allocating resources to changing $\mathbb{C}$ or $\mathbb{S}$. Learning would then be represented as a process of testing different allocation strategies, seeing if a strategy improves effectiveness and/or increases the production of resources, and then updating the priors accordingly. Modelling resource allocation in such a way would allow us to both investigate the feasibility of reaching a balanced allocation as well as observe if skewed allocation strategies converge on sustainable co-evolutionary dynamics.

A fourth and final assumption is that societies do not interact. Indeed, in our model, each simulation run had only one evolving society. To some extent, this assumption could be taken to represent the entirety of human civilization, but it does overlook the richer dynamics of modelling societies as entities that interact with one another through competition (e.g., warfare) and cooperation (e.g., trade) over resources [120]. A more ambitious extension in this vein would model multiple societies via a fission-fusion process [121,122]. Societies could now compete and cooperate, but also proliferate over time by branching off into distinct ones (fission), homogenise by integrating into larger civilizations (fusion), as well as collapse and disappear through processes already modelled in this paper (i.e., absorbing barriers). Future research is now well-positioned to link each of these processes to the ability of societies to develop effective and complex cultural systems, to exploit and expand their search spaces, and to produce and allocate resources.

## 5 Conclusion

Any theory of human cultural evolution, which seeks to explain its cumulative, adaptive and open-ended nature, needs to account for the relationship between cultural systems ($\mathbb{C}$) and the changing needs, problems and goals of societies (what we termed search spaces, $\mathbb{S}$). Our model tentatively provides the stepping stones for such a theory by positing that $\mathbb{C}$ and $\mathbb{S}$ are subject to evolutionary dynamics and capable of co-evolving with one another. Crucially, we formulated the relationship between $\mathbb{C}$ and $\mathbb{S}$ in terms of their ability to produce resources. This sets up a non-trivial challenge for the dynamics of our model: societies need to overcome the resource costs of maintaining complex cultural systems by finding effective outcomes *and* through continually increasing their access to resources by expanding the search space.

By controlling the extent to which the co-evolutionary dynamics are stochastic or deterministic, our model arrived at three general conclusions. First, sustaining long-term co-evolution is difficult when societies need to produce resources, with absorbing barriers representing a hard constraint on the emergence of open-ended growth. Second, open-ended growth is possible so long as stochastic factors keep the dynamics in a far-from-equilibrium state, and deterministic factors increasingly adopt expansionary changes. Third, regions of the parameter space most conducive to open-ended growth are those characterised by significant selection pressures on both $\mathbb{C}$ and $\mathbb{S}$. Understanding long-term cultural evolution is therefore going to require a fully articulated theory of how cultural systems and search spaces co-evolve over time.